\documentclass[final,3p,times,twocolumn]{elsarticle}

\usepackage{amssymb}
\usepackage{amsmath,graphicx,multirow, color,subfigure,floatrow}
\usepackage{slashbox}

\begin{document}

\begin{frontmatter}



\title{Fusion Based Holistic Road Scene Understanding}
	\author[au1]{Wenqi Huang}
	\author[au1]{Xiaojin Gong\corref{cor1}}
	\ead{gongxj@zju.edu.cn}
  \cortext[cor1]{Corresponding author}
  \address{Dept. of Information Science and Electronic Engineering, Zhejiang University, \\ Hangzhou 310027, Zhejiang, P.R. China}


\begin{abstract}
This paper addresses the problem of holistic road scene understanding based on the integration of visual and range data. To achieve the grand goal, we propose an approach that jointly tackles object-level image segmentation and semantic region labeling within a conditional random field (CRF) framework. Specifically, we first generate semantic object hypotheses by clustering 3D points, learning their prior appearance models, and using a deep learning method for reasoning their semantic categories. The learned priors, together with spatial and geometric contexts, are incorporated in CRF. With this formulation, visual and range data are fused thoroughly, and moreover, the coupled segmentation and semantic labeling problem can be inferred via Graph Cuts. Our approach is validated on the challenging KITTI dataset that contains diverse complicated road scenarios. Both quantitative and qualitative evaluations demonstrate its effectiveness.
\end{abstract}


\end{frontmatter}



\section{Introduction}
\label{sec:introduction}
\begin{figure*}[htbp]\tiny
\centering
  \subfigure{
  \begin{minipage}[t]{0.9\linewidth}
  \includegraphics [width=1\textwidth]{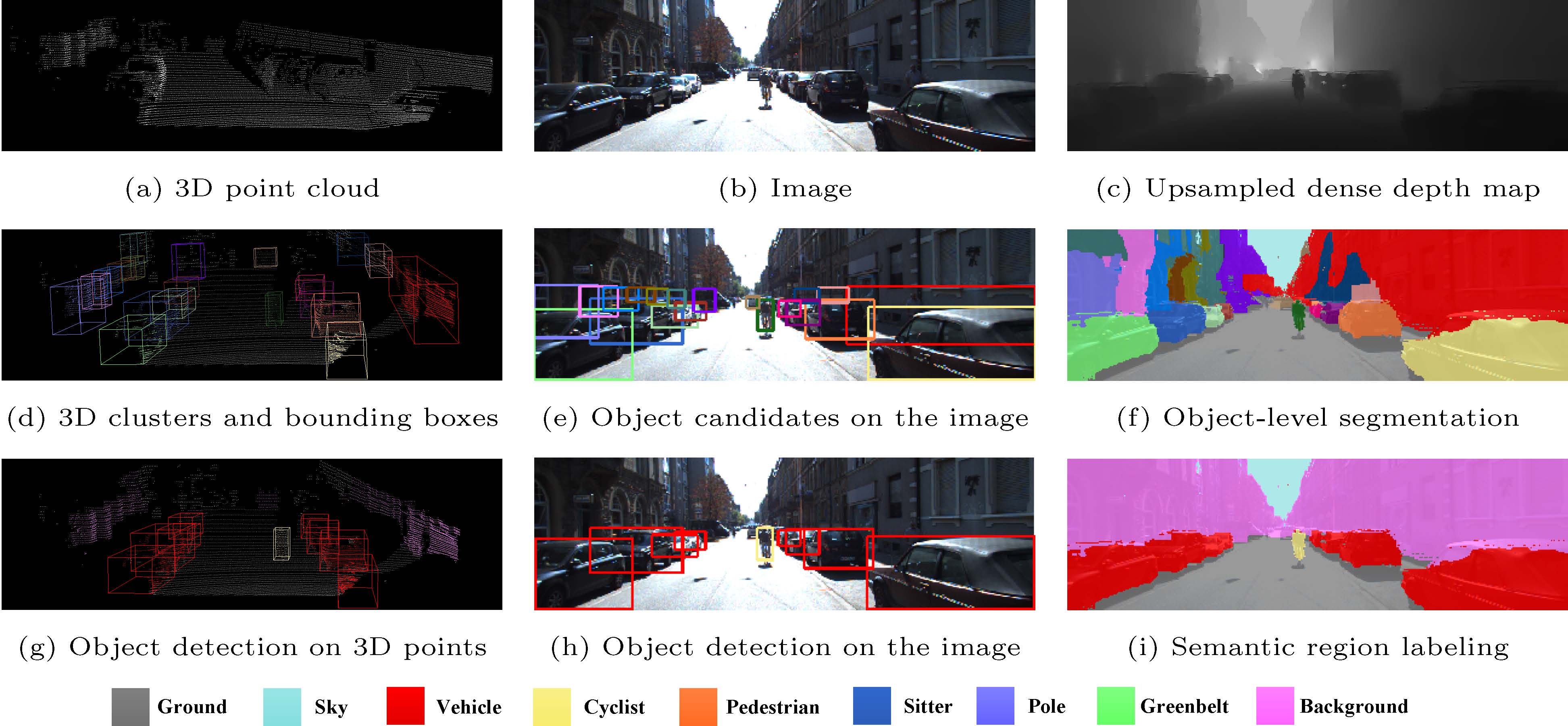}
  \end{minipage}
  }
  \vspace{-1em}
  \caption{An overview of the tasks achieved in this work. Given an aligned 3D point cloud (a) and a color image (b), we first obtain a dense depth map (c) by a guided upsampling technique. Then, the 3D point cloud is clustered to generate object hypotheses (d). The bounding cuboids are projected onto the image to get object candidates (e). Both object-level image segmentation (f) and semantic region labeling (i) are obtained simultaneously by our proposed approach. From them, we directly get the object detection results on the image (h) and on the point cloud (g). Note that the colors in the second row have no semantic meaning. Different colors denote different object instances. The colors in the third row represent the corresponding semantic categories, as shown in the legend.}
  \label{figure:Overview}
\end{figure*}

Road scene understanding plays an important role in various computer vision applications, ranging from autonomous driving to urban modeling. It commonly involves multiple tasks, such as drivable road surface detection~\cite{Alvarez2010, Huang_ICIP2013}, pedestrian and vehicle detection~\cite{Benenson_CVPR2012,YunFu2014,Nguyen2013,Yangqing2009}, semantic region labeling~\cite{Huanhuan2010,Levinkov_ICCV2013, Guo_ITSC2011, Alvarez_2012,Huang_2014,Cheng_PR2010}, geometric context reasoning~\cite{Jung_PR2012,Matzen_ICCV2013}, and so on. Each individual task is notoriously difficult due to the complexity of natural scenarios. As in the typical example presented in Fig.~\ref{figure:Overview} (b), a road scene may contain severe lighting variation and a cluttered roadside background, together with variant numbers of vehicles and pedestrians. These challenges have led to a large amount of studies on tackling each problem.

Most existing work addresses the above-mentioned tasks individually. However, we can observe that these problems are coupled. For example, semantic region labeling should be easier if we know where the ground plane and moving objects are. Likewise, geometric context helps to detect objects and label regions. These observations inspire our research here. In order to take advantage of the benefits from such correlations, this paper proposes to solve the problems jointly. In addition, considering that cameras and ranging sensors are often used conjunctively on today's autonomous vehicles, we build our work upon the fusion of visual and range data.

Specifically, this paper proposes a holistic approach that exploits appearance, geometry and contextual information to jointly tackle object-level image segmentation and semantic region labeling, from which it is straightforward to locate drivable road surfaces and moving objects in both images and 3D point clouds, as illustrated in Fig.~\ref{figure:Overview} (f)-(i). Holistic road scene understanding is consequently achieved, providing robots with a deeper understanding of the whole scene.

The proposed approach distinguishes itself from other holistic scene understanding techniques in a couple of aspects. First, our approach generates semantic object hypotheses by simply clustering a 3D point cloud into object candidates, learning their prior Gaussian mixture models (GMMs), and using a deep learning method to reason their semantic categories. This procedure does not involve sophisticated feature extraction and requires almost no tedious pixel-wise hand labeling. Second, we perform bimodal data fusion on multiple stages, hierarchically, from image guided depth map upsampling to RGB-D image patch based object classification and holistic inference in a conditional random field (CRF). Thus, both visual and range information are thoroughly utilized. Last but not least, to the best of our knowledge, this research is one of the first studies working on holistic road scene understanding. The effectiveness of our approach is validated on the challenging KITTI dataset~\cite{Geiger_CVPR2012}.

The remainder of this paper is organized as follows. In Section~\ref{sec:related_work}, we make a brief review of both fusion-based and holistic oriented scene understanding techniques. Section~\ref{section: Hypothesis} introduces the method of generating semantic object hypotheses. The proposed holistic CRF framework, which incorporates the learned priors, together with lidar point pivoted hard constraints and geometric context, to jointly solve the problems, is presented in Section~\ref{sec:CRF}. Experiments are demonstrated in Section~\ref{sec:Experiments}, followed by a conclusion in Section~\ref{sec:Conclusions}.

\section{Related Works}
\label{sec:related_work}
There is a huge body of work related to our problem in that it encompasses multiple extensively studied tasks. In this section, we focus our attention on the two most relevant aspects, which are fusion-based and holistic scene understanding. The former emphasizes the fusion of multi-modal data for the tasks and the latter aims to solve multiple tasks jointly. 

\subsection{Fusion Based Scene Understanding}
With the advent of ranging sensors, nowadays, it is quite convenient for us to capture synchronized range and visual data. Such convenience has motivated a great number of studies on fusing these two modalities for tasks towards scene understanding. In contrast to a camera- or lidar-only scheme, fusion dramatically increases accuracy and robustness in various applications.

Generally speaking, fusion is often conducted at feature or decision level. The feature-level methods fuse two modalities via extracting both appearance and geometric features and concatenating them together for the succeeding process. Particularly, these methods first segment RGB-D data into superpixels~\cite{Ren_CVPR2012}, divide a colored 3D point mesh into spatially adjacent regions~\cite{Strom_IROS2010, Valentin_CVPR2013}, or map both pixels and 3D points into cells~\cite{Haselich_ECMR2011, Laible_AMS2012}. Then, sophisticated appearance features, such as texton~\cite{Shotton2006}, SIFT and HOG~\cite{Silberman2011}, and kernel descriptors~\cite{Ren_CVPR2012}, as well as geometric features, such as surface normal, angular moments, and average height, are extracted from each unit for the tasks of object detection, 3D point segmentation~\cite{Strom_IROS2010, Schoenberg_IROS2010}, terrain classification~\cite{Haselich_ECMR2011, Laible_AMS2012}, semantic 3D modeling~\cite{Valentin_CVPR2013}, and scene parsing~\cite{ Guo_ITSC2011,Zhao_FUSION2012}. Among these studies, RGB-D data oriented work is mostly limited to indoor scene parsing because a great portion of such data are obtained by Kinect-like sensors (although they can also be obtained by upsampling lidar data~\cite{Diebel_NIPS2006}). In contrast, 3D point clouds are collected by lidar so that they are more suitable for outdoor applications.

In contrast to feature-level fusion, a decision-level method analyzes each modality individually and then combines the analysis results through a fusion scheme. For instance, Zhao et al.~\cite{Zhao_FUSION2012} utilize the fuzzy logic inference framework to combine the classification results of lidar data and that of images for scene parsing.

Other than the two above-mentioned separate fusion schemes, the use of deep learning, which is a powerful architecture merging both feature- and decision-level fusion into a whole, surged recently. It learns both feature representation and classification simultaneously to solve tasks such as RGB-D based object recognition~\cite{Socher2012} and demonstrates promising results.

In contrast, our approach integrates visual and range information on multiple stages. More specifically, low-level fusion is first conducted to produce dense depth maps by using an image guided depth upsampling technique~\cite{Liu2013} previously proposed by us. The obtained RGB-D image patches are fed into a deep learning method as well to reason semantic categories. Finally, in the proposed holistic conditional random field framework, besides the learned appearance and geometric priors, lidar points are integrated as hard constraints to guide image segmentation. Therefore, our fusion is conducted in a hierarchical way, which thoroughly makes use of the bimodal information.

\subsection{Holistic Scene Understanding}
While substantial progress has been made in numerous computer vision tasks over the last few decades, most previous works tackled each particular problem isolatedly. In recent years, however, more researchers have started to exploit the dependencies between different tasks and attempted to solve two or more problems jointly. For example, Bleyer et al.~\cite{Bleyer_CVPR2011}, Ladicky et al.~\cite{Ladicky_IJCV2012} and Hane et al.~\cite{Hane_CVPR2013} combine stereo reconstruction with object segmentation to improve the performance of both. The problems of classification and segmentation are also simultaneously addressed in~\cite{Gonfaus_CVPR2010}. In light of these successes, researchers have stepped further toward achieving the grand goal of holistic scene understanding~\cite{Heitz_NIPS2008, Dahua_ICCV2013,Li_PAMI2012,Yao_CVPR2012}.

Holistic scene understanding aims to fully interpret a scene by jointly solving the tasks of image segmentation, object detection, 3D reconstruction, scene classification, etc. To achieve this target, a critical problem that we face is how to infer mutual information between the tasks. Here, we roughly categorize the inference techniques into two groups. One develops a general framework, such as Cascaded Classification Models (CCM)~\cite{Heitz_NIPS2008} and feedback enabled CCMs~\cite{Li_PAMI2012}, to combine different tasks. These techniques treat the components of each task as black boxes. They rely upon complicated inference algorithms so it is hard to incorporate potentials specific to some particular problems~\cite{Yao_CVPR2012}.

A more extensive method is formulating a joint problem as an inference within a Markov or conditional random field (CRF) framework~\cite{Bleyer_CVPR2011,Ladicky_IJCV2012,Gonfaus_CVPR2010,Dahua_ICCV2013, Yao_CVPR2012, Ladicky_ECCV2010, Tighe_ICCV2011}. Each node in the graph represents a segmentation or category label associated with a pixel, superpixel or 3D point. Potentials encode unitary information and pairwise or high-order relations of inter- or intra-tasks. Inference within the random field is done by either a message-passing approach~\cite{Yao_CVPR2012}, fusion moves~\cite{Bleyer_CVPR2011}, or more efficient Graph Cuts algorithms~\cite{Ladicky_IJCV2012, Gonfaus_CVPR2010, Ladicky_ECCV2010, Boykov01} if energy functions satisfy submodularity restriction. In summary, the differences among all CRF-based works rely on the problems to be solved, the construction of the graphical models, the incorporated priors, and the inference techniques.

Our work follows the second line in order to thoroughly exploit the priors specific to road scenes and hierarchically fuse the bimodal data. The proposed holistic CRF graphical model is used for us to jointly solve object-level image segmentation and semantic region labeling problems. Our CRF encodes the priors learned from the bimodal data, together with lidar point pivoted hard constraints and geometric context, in the unary potentials. Meanwhile, pairwise potentials exploit the spatial dependencies in each task, as well as the coherency between the two tasks. All designed unary and pairwise potentials meet the submodularity restriction, so that Graph Cuts can be used for efficient inference.

\section{Semantic Object Hypotheses Generation}
\label{section: Hypothesis}
Before integrating all information within a CRF framework, the first stage for us is to generate initial object hypotheses, learn their prior models, and reason their semantic categories. Considering that geometric information is more reliable than visual cues for discovering objects, we start from partitioning a 3D point cloud into clusters to obtain object hypotheses. Once we get the clustered points, their registered pixels, which are also referred to as \textit{seeds}, are taken to build prior models of the objects. Moreover, each RGB-D image patch that is registered to the bounding cuboid of a 3D cluster is fed into a convolutional recursive neuron network (CRNN)~\cite{Socher2012} to determine its semantic category. The details of each step are stated below.

\subsection{Data Preprocessing}
The data we process are aligned image-lidar pairs that are, respectively, collected by a camera and a lidar mounted on a vehicle~\cite{Geiger_CVPR2012}. When the intrinsic and extrinsic parameters of both sensors are calibrated, it is handy for us to register a 3D point set and an image to each other. By registration, we obtain a sparse depth map, in which the seeds are assigned with corresponding depth values and the remainder is of no depth information. For the convenience of the subsequent processes, the sparse depth map is upsampled by a guided depth enhancement technique~\cite{Liu2013}, which generates a dense depth map via integrating the sparse one with a color image. An example result is illustrated in Fig.~\ref{figure:Overview}(c).

\subsection{Generating Object Hypotheses}
As pointed out by Douillard et al.~\cite{Douillard_ICRA2011}, the ground extraction significantly improves clustering performance. Therefore, before 3D point clustering, we first estimate the ground plane. The ground is commonly the dominant plane in most road scenes. We therefore use the Random Sample Consensus (RANSAC) algorithm~\cite{Fischler1981} to estimate it. However, in scenarios such as a narrow street with buildings on both sides, the estimated dominant plane may lie on a wall of the buildings. In order to avoid such a mistake, we define a rough range for height according to where the lidar is equipped on the vehicle. Only the 3D points within the range are taken into consideration for ground plane estimation. 

After detecting the ground plane, we leave out the corresponding points and use a simple but effective Euclidean clustering method to partition the remainder to generate object candidates. This method~\cite{PCL2013} is based on the nearest neighbor scheme. It is implemented with a kd-tree data structure and therefore is quite efficient. Moreover, this approach produces a set of object clusters well, especially for separated objects on the road.

Note that our clustering is performed on the original sparse 3D lidar points, instead of the denser points reconstructed from the upsampled dense map. The reason is that the upsampling techniques are prone to generate artifacts, especially on the places near object boundaries and in large invalid regions, leading to errors that might be propagated to later stages.

\subsection{Learning Object Priors}
\label{sec:obj_prior}
Once the ground and other object clusters are produced, the corresponding seeds are taken as samples to learn their prior models. In our work, we only take the RGB color and 3D location of each seed as our feature. No other sophisticated features are considered. Therefore, for each object instance, a Gaussian mixture model (GMM) of the 6D feature $(R, G, B, X, Y, Z)$ is built. It needs to be mentioned that a different means is taken for building the sky model. Since there is no way to sample the sky from lidar data, sky regions in a set of images are manually labeled to learn a color GMM for the sky.

\subsection{Reasoning Semantic Categories}
This step is to determine the semantic category for each image patch registered to a 3D cluster. In order to avoid the complicated feature extraction step, we simply apply a deep learning method here. More specifically, a convolutional recursive neuron network (CRNN)~\cite{Socher2012} is adopted, which takes a RGB-D image patch as input. Within the CRNN, a convolutional neural network (CNN) layer with weights trained by k-means clustering is first used to extract low level features from the patch. The resulting feature maps are then connected to several recursive neural networks (RNN) to get higher-order combinational features. The weights of the RNNs are randomly assigned, which is very efficient and has shown to be good enough. Finally, the RNNs' outputs are fed into a softmax classifier for recognition. The CRNN associates each image patch with a set of scores, indicating the confidence of it being a specific category.

\section{Holistic CRF Model}
\label{sec:CRF}
\begin{figure*}[htbp]
\centering
\includegraphics [width=0.9\textwidth]{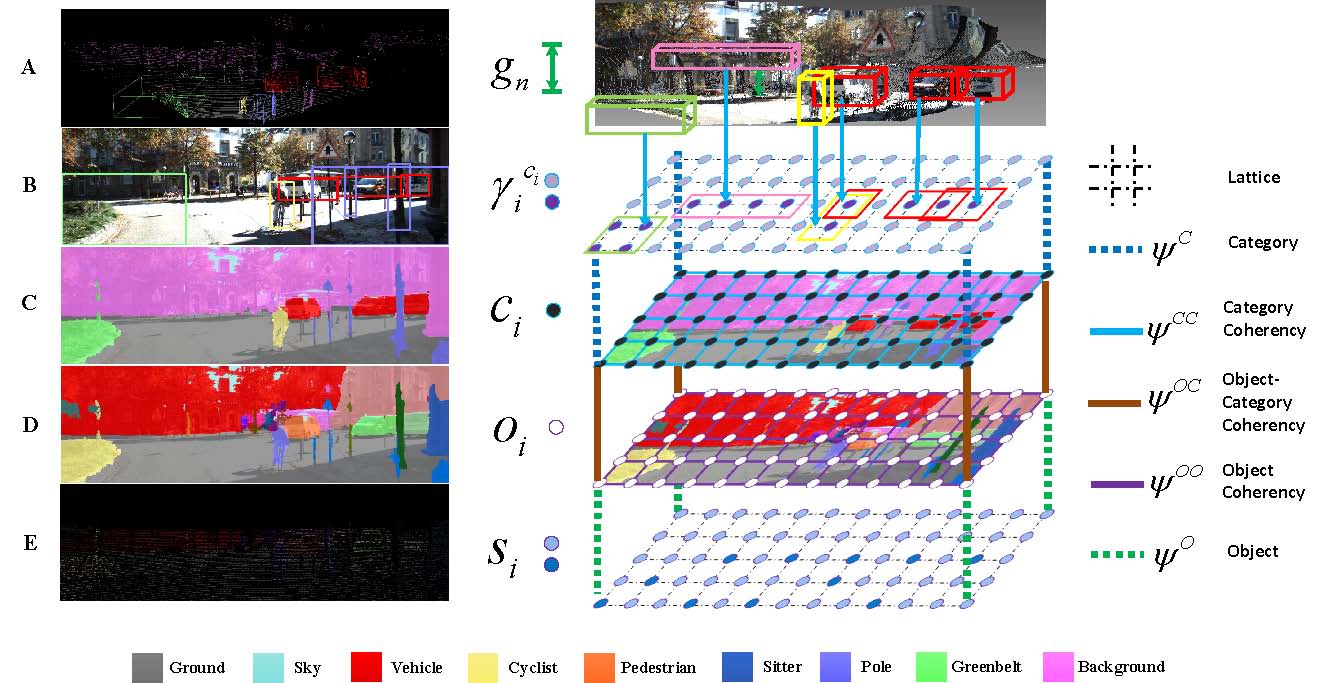}

  \caption{An illustration of the proposed CRF model. It consists of two hidden layers of random variables associated with each pixel, one ($\{o_i\}$) for object-level segmentation and the other ($\{c_i\}$) for semantic region labeling. The CRF model integrates the observed features $\mathbf{f}_i = (R_i, G_i, B_i, X_i, Y_i, Z_i)$, together with the seeds $\{s_i\}$, pivoted hard constraints and the geometric contextual information $\{g_n\}$ to infer the joint problem. Specifically, the deep blue points on the layer of $\{s_i\}$ indicate the sparse seeds and the deep purple points on the layer of $\{\gamma_i^{c_i}\}$ indicate the points in a patch. The images on the left side are category recognition on 3D points (image A), recognition on image (image B), category labeling result (image C), object segmentation result (image D) and object seeds (image E).  Note that the colors on image D have no semantic meaning and different colors denote different objects. The colors on image C represent the corresponding semantic categories, as shown in the legend.}
  \label{figure:FlowChart}
\end{figure*}

In this section, we formulate road scene understanding as a labeling problem, which associates each pixel with two types of labels: one indicates an object instance that the pixel belongs to and the other tells its semantic category. To this end, we construct a holistic CRF model consisting of two hidden layers. The model also integrates observed features of the pixels, together with the 3D lidar points and geometric contextual information to boost the accuracy of both object-level segmentation and semantic region labeling. Fig.~\ref{figure:FlowChart} illustrates our constructed model.

Formally, when an image $\mathbf{I}$ is given, we construct a graph $\mathcal{G} =\left\langle \mathcal{V}, \mathcal{E}\right\rangle$. Here, the vertex set $\mathcal{V} = \{\mathcal{V}_{\mathcal{O}}, \mathcal{V}_{\mathcal{C}}\}$ consists of two sets of random variables and the edge set $\mathcal{E} = \{\mathcal{E}_{\mathcal{O}\mathcal{O}}, \mathcal{E}_{\mathcal{C}\mathcal{C}}, \mathcal{E}_{\mathcal{O}\mathcal{C}}\}$ contains three types of edges. More specifically, a random variable $o_i \in \mathcal{V}_{\mathcal{O}}$ is associated with the $i$-th pixel and takes a value from $\{0, \cdots, O+1\}$ to represent the $o_i$-th object label, in which $O$ is the total number of object hypotheses generated in Sec.~\ref{sec:obj_prior}, $0$ is for the ground and $O+1$ is for the sky. Likewise, a random variable $c_i \in \mathcal{V}_{\mathcal{C}}$ takes a value from $\{1,2, \cdots, C\}$ to indicate its category, where $C$ is the total number of semantic categories. With such a graphical model, an optimal solution of joint object-level segmentation and semantic region labeling is obtained by maximizing the following probability:
\begin{equation}
\label{eq:prob}
\begin{split}
&p(\mathbf{o}, \mathbf{c}) = \frac{1}{Z} \exp \left( \lambda_1 \sum_{i=1}^N \psi^{O}(o_i) + \lambda_2 \sum_{i=1}^N \psi^C(c_i) \right.\\
                          &\ \ \ + \lambda_3 \sum_{i=1}^N \sum_{e_{ij} \in \mathcal{E}_{\mathcal{O}\mathcal{O}}} \psi^{OO}(o_i, o_j) + \lambda_4 \sum_{i=1}^N \sum_{e_{ij} \in \mathcal{E}_{\mathcal{C}\mathcal{C}}} \psi^{CC}(c_i, c_j) \\
                          &\ \ \ + \left. \lambda_5 \sum_{i=1}^N \sum_{e_{ii} \in \mathcal{E}_{\mathcal{O}\mathcal{C}}}\psi^{OC}(o_i, c_i)\right),
\end{split}
\end{equation}
where $Z$ is the partition function. In addition, there are five types of potentials. $\psi^{O}(o_i)$ and $\psi^C(c_i)$ are two unary potentials associated with the object label and the category label, respectively. $\psi^{OO}(o_i, o_j)$ is a pairwise potential exploiting the dependency of neighboring object labels. $\psi^{CC}(c_i, c_j)$ is also a pairwise potential investigating the dependency of category labels. $\psi^{OC}(o_i, c_i)$ investigates the mutual information between object labels and category labels, and $\lambda_1, ..., \lambda_5$ are scaling factors. The details of each potential are explained below. With appropriate design, this graphical model can be inferred with the efficient Graph Cuts algorithm~\cite{Boykov01}.

\subsection{Object Potential}
\label{sec:ObjectPotential}
The object potential evaluates the confidence for a pixel to be labeled as the $o_i$-th object. Commonly, it is designed in terms of the likelihoods, as follows~\cite{Boykov01,Rother2004}:
\begin{equation}
\label{eq:Data}
\psi^{O}(o_i) = - \ln p(\mathbf{f}_i | \Theta_{o_i}),
\end{equation}
where $\mathbf{f}_i = (R_i, G_i, B_i, X_i, Y_i, Z_i)$ is the feature vector associated with the $i$-th pixel, $\Theta_{o_i}$ denotes the parameters of the $o_i$-th object's GMM that we have learned in Sec.~\ref{sec:obj_prior}, and $p(\mathbf{f}_i | \Theta_{o_i})$ is the likelihood.

The above-defined likelihood potential is sensitive when two objects share similar features. For instance, strong shadows on the ground and bushes nearby are prone to be labeled as the same object by mistake. In contrast, 3D point clustering performs better; at least it is invariant to illumination change. Therefore, we place high confidence~\cite{Rother2004} on the seeds. Let us denote the entire set of seeds by $\mathcal{S}$, and the set of seeds belonging to the $o$-th object by $\mathcal{S}_o$. Then, the object potential is placed with hard constraints (HC) and defined by
\begin{equation} \label{eq:Fusion1}
\psi^{O}(o_i) =\left\{ \begin{array}{ll}
\alpha_o & \textrm{$i \in \mathcal{S}_{o_i}$}\\
\beta_o & \textrm{$i \in \mathcal{S} / \mathcal{S}_{o_i}$}\\
- \ln p(\mathbf{f}_i | \Theta_{l_i})  & \textrm{otherwise,}
\end{array} \right.
\end{equation}
where $\alpha_o$ is a small positive value and $\beta_o$ is a large positive value, which are experimentally set to force the constraints. With these hard constraints, the labels of the registered pixels are forced to be consistent with the point clustering results.

\subsection{Category Potential}
The category potential indicates the confidence for a pixel to be the $c_i$-th category. This potential incorporates the classification result obtained by the CRNN together with the learned prior models and geometric contextual information for better reasoning.

Specifically, for the purpose of simplicity, let us first divide the semantic categories into three groups: $\mathcal{C}_{SG}$, $\mathcal{C}_{B}$, and $\mathcal{C}_{O}$. $\mathcal{C}_{SG}$ stands for either the ground or the sky category, $\mathcal{C}_{B}$ contains the background category and $\mathcal{C}_{O}$ is for the remaining categories, such as pedestrians, vehicles, etc. The latter two are recognized by the CRNN. Therefore, we define a confidence score $f(P_k, c)$ for an image patch $P_k$ to be the category $c$, which is
\begin{equation}
\label{eq:f}
f(P_k, c)  = \left\{\begin{array}{ll}
1 & \ \ c \in \mathcal{C}_{SG} \cap k \in \{0,O+1\} \\
0 & \ \ c \in \mathcal{C}_{SG} \cap k \in \{1,2,...,O\} \\
s(P_k, c) & \ \ c \in \mathcal{C}_{B} \\
s(P_k, c) \cdot g(P_k) & \ \ c \in \mathcal{C}_{O},
\end{array} \right.
\end{equation}
where $k \in \{1, \cdots O\}$ denotes the $k$-th object hypotheses, $k = 0$ for the ground and $k = O+1$ for the sky, as before. Note that, there is no patch for the ground and sky. For a uniform formulation, we define the patch of ground, denoted as $P_0$, as the part under the horizon line~\cite{Huang_ICIP2013} of the image and the patch of the sky, $P_{O+1}$, as the rest of the image. $s(P_k, c)$ is the score obtained by the CRNN. $g(P_k)$ is a term introducing geometric properties. Although more complicated geometric relations can be taken into account, here we only investigate a quite straightforward observation. That is, except the ground, the sky, and the background, all other objects must lie on the ground. Therefore, this constraint is designed to be
\begin{equation}
\label{eq:G}
g(P_k)  = \left\{\begin{array}{ll}
1 & \text{bottom\_height}(P_k) < T_h \\
0 & \textrm{otherwise}.
\end{array} \right.
\end{equation}
Here, $\text{bottom\_height}(P_k)$ denotes the bottom height of the corresponding object cuboid, which should be lower than a threshold $T_h$.

\begin{figure*}[htbp]
\centering
\includegraphics [width=0.8\textwidth]{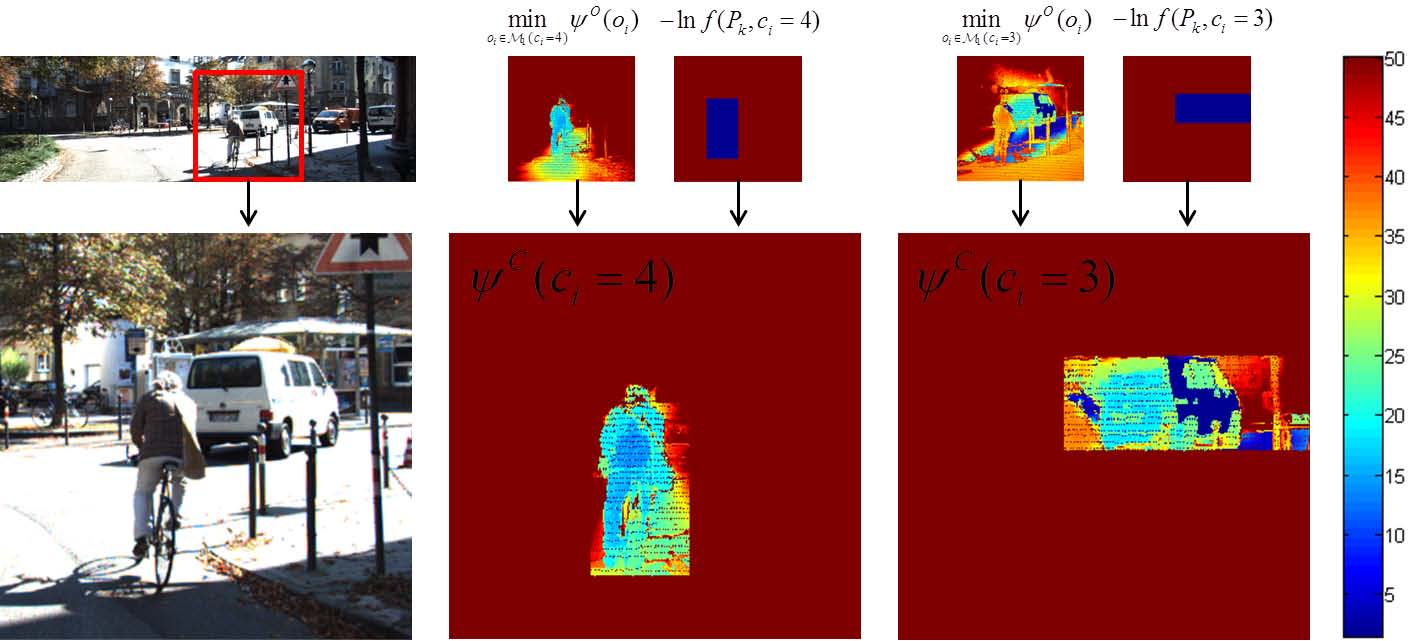}
  \caption{An illustration of category potential. Here we use $c_i = 3$ to represent the category of vehicle and $c_i=4$ for cyclist.}
  \label{figure:Potential}
\end{figure*}

Upon these, we define our category potential as below:
\begin{equation}
\label{eq:CP}
\psi^C(c_i)  = \left\{\begin{array}{ll}
\min\Big(- \ln f(P_k, c_i) + \min\limits_{o_i \in \mathcal{M}_1(c_i)}\psi^O(o_i), \ \alpha_c \Big) & i \in P_k\\
\alpha_c & \textrm{otherwise}.
\end{array} \right.
\end{equation}
Here, $\mathcal{M}_1(c_i)$ denotes the set of object instances that are identified as the $c_i$-th category; $\alpha_c$ is a large positive value assigned for the pixels that are not falling into any object patches. The reason to combine the category recognition confidence $f(P_k, c_i)$ together with the object-level segmentation confidence $\psi^O(o_i)$ is for obtaining semantic labeling results with better object boundaries. An illustration of this term is presented in Fig.~\ref{figure:Potential}.

\subsection{Object Coherency Potential}
The object coherency potential exploits the dependence between neighbors. It encourages two neighboring pixels to take the same object label if their associated features are similar to each other. This potential can smooth out isolated labels, leading to piecewise coherent results.

Specifically, for a pixel $v_i$ and each of its 4-connected neighboring pixels $v_j$, this potential is defined as
\begin{equation} \label{eq:Smooth1}
\psi^{OO}(o_i, o_j) = \exp\left(-\frac{||\mathbf{f}_i - \mathbf{f}_j||_2^2}{\sigma^2}\right) \cdot T(o_i \neq o_j),
\end{equation}
where $||\mathbf{f}_i - \mathbf{f}_j||_2$ is the $L_2$ norm of the difference between the features $\mathbf{f}_i$ and $\mathbf{f}_j$. $T()$ is an indicator, whose value is 1 when its parameter is true and 0 otherwise. This term indicates that the more similar the features are, the more likely that the two pixels belong to the same object.

\subsection{Category Coherency Potential}
The category coherency potential encourages neighboring pixels to take the same category label. Likewise, it is defined by
\begin{equation} \label{eq:Smooth2}
\psi^{CC}(c_i, c_j) = \exp\left(-\frac{||\mathbf{f}_i - \mathbf{f}_j||_2^2}{\sigma^2}\right) \cdot T(c_i \neq c_j).
\end{equation}

\subsection{Object-Category Coherency Potential}
\label{sec:O_C}
This potential is proposed to exploit the dependency between object and category labels of the same pixel. More specifically, the category label of a pixel should be the same as the recognition result of the object that the pixel belongs to. Therefore, it is designed as
\begin{equation} \label{eq:CoherencyPotential}
\psi^{OC}(o_i,c_i) = T\Big(c_i \neq \mathcal{M}_2(o_i)\Big),
\end{equation}
where $\mathcal{M}_2(o_i)$ is a function determining the category that an object instance belongs to, which is defined as:
\begin{equation} \label{eq:correspondence2}
\mathcal{M}_2(o_i) =\arg\max\limits_{c}f(P_k, c), \ \ \ k=o_i.
\end{equation}

\section{Experiments}
\label{sec:Experiments}

\subsection{KITTI Dataset}
In order to validate the proposed approach, we have conducted a series of experiments on the KITTI vision benchmark suite~\cite{Geiger_CVPR2012}, which provides us with numerous color images and 3D point clouds. The data are captured by a PointGrey Elea2 video camera and a Velodyne HDL-64E 3D lidar that are jointly mounted on a vehicle. Each image is in the resolution of $1242 \times 375$, and a 3D point cloud is of $100,000$ points or so, which covers a $360^o$ field of view (FOV). But only the points falling within the camera's FOV are taken into consideration. The two modalities are registered to each other according to the sensors' parameters provided on KITTI's website.

Experiments are conducted on the 'City', 'Residential', and 'Road' datasets, which contain a variety of complex scenarios on urban and highway roads, with the presence of vehicles, cyclists, pedestrians and other objects. The total number of images is 18529, among which 13765 images are randomly selected for the CRNN and the remaining 4764 images are used for evaluation. The details of the evaluation are stated below.

\subsection{Evaluation of CRNN}
The step of semantic reasoning via the CRNN is critical for our final results. Therefore, we first evaluate its performance. The input of the CRNN is an image patch obtained in the way introduced in Sec.~\ref{section: Hypothesis}. More specifically, we use the nearest neighbor clustering algorithm in the Point Cloud Library (PCL)~\cite{Rusu_ICRA2011} to generate initial object hypotheses. The produced clusters that have a very small number of faraway points are discarded for robustness. Then, the image patches registered to these clustered 3D points are fed into the CRNN as inputs.

Each patch is resized to $67 \times 67$. In the CRNN~\cite{Socher2012}, we set the size of a CNN filter to $8 \times 8$ and the number of filters is 128. Pre-training for CNN filters is performed by k-means clustering on 300,000 patches, randomly sampled from our training set. Average pooling is performed with pooling regions of size 8 and stride size 2 to produce 128 feature maps of the size of $27 \times 27$. The RNN receptive field size is set to $3\times 3$, by which each feature map is recursively reduced to size $9 \times 9$, to $3 \times 3$, and finally to $1\times1$. Through four RNNs, the final feature for classification is $128 \times 4$.

We manually label all the patches extracted from $13765$ images into seven object categories. The categories and their corresponding patch numbers are listed in Table~\ref{tab:ObjectSample}. In each category, we randomly select $70\%$ patches for the CRNN training and the rest for the CRNN testing. We also horizontally flip the patches in the 'Cyclist', 'Pedestrian', and 'Sitter' categories in order to double their training samples.

\begin{table*}[htbp]\tiny\centering
\begin{tabular}{|c||c|c|c|c|c|c|c|}
\hline
Object Category & Vehicle & Cyclist & Pedestrian &Sitter & Pole & Greenbelt & Roadside \\
\hline
\hline
Sample Number  &  $15020$ & $789$ & $567$ & $25$ & $2590$ & $3498$ & $38459$ \\
\hline
\end{tabular}
\caption{Object categories and the corresponding sample numbers.}
\label{tab:ObjectSample}
\end{table*}

In this section, a set of comparative experiments are designed in order to investigate the performance of the CRNN with different input configurations. For instance, we compare the performance of the CRNN when using RGBD patches versus that of using RGB only. Moreover, although rectangular patches are fed into the CRNN, our algorithm is actually able to extract object regions. Therefore, we also compare the performance for patches with and without masks. The average recognition accuracy of each configuration is shown in Table~\ref{tab:CRNN1}. It shows that the CRNN performs the best when depth information is considered and the background is masked out. 

\begin{table*}[htbp]\tiny\centering
\begin{tabular}{|c|c|c|c|c|}
\hline
 \multirow{2}*{Configuration}& \multicolumn{2}{|c|}{Unmasked} & \multicolumn{2}{|c|}{Masked} \\
\cline{2-5}
& RGB & RGBD & RGB & RGBD  \\
\hline
\hline
Average Accuracy& $87.01\%$ &$ 87.87\%$ & $88.05\%$ & $\mathbf{89.32\%}$ \\
\hline
\end{tabular}
\caption{Recognition accuracy of CRNN.}
\label{tab:CRNN1}
\end{table*}

\begin{figure*}[htbp]
\centering\tiny
  \subfigure{
  \begin{minipage}[t]{0.35\linewidth}
  \centering
   \includegraphics [width=1\textwidth]{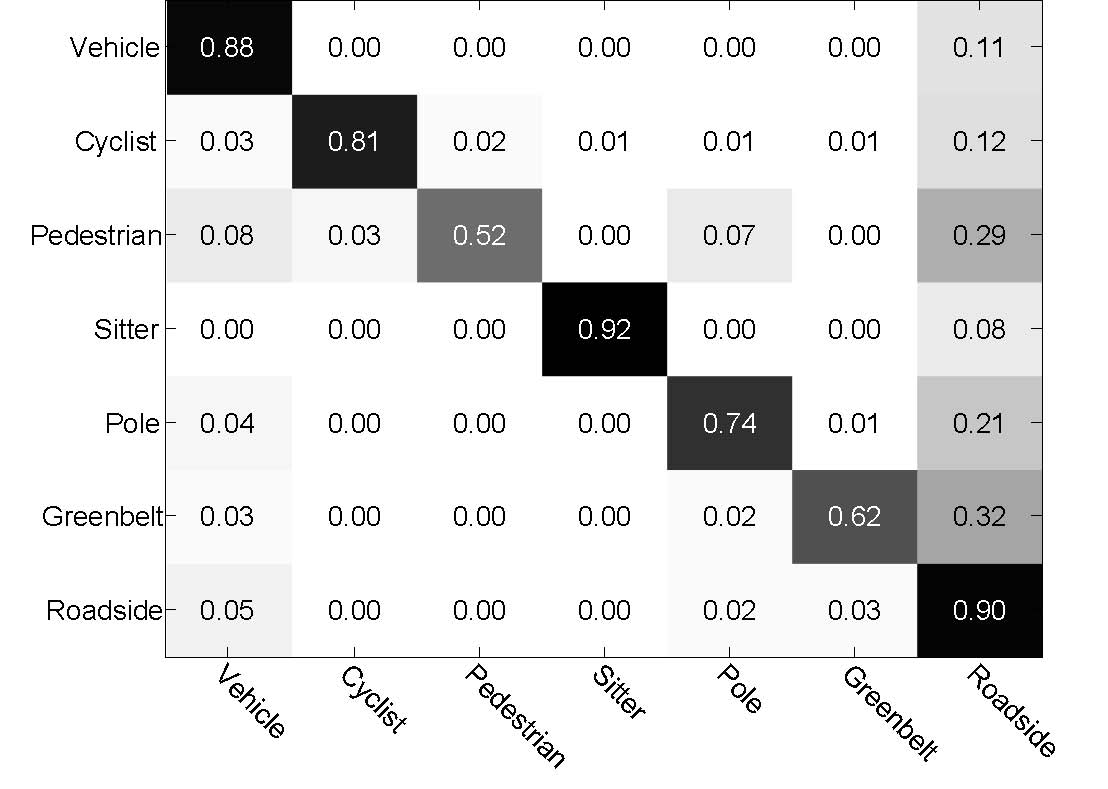}
   \centerline{(a) Unmasked RGB}\medskip
  \end{minipage}
  \hfill
  }
  \subfigure{
  \begin{minipage}[t]{0.35\linewidth}
   \includegraphics [width=1\textwidth]{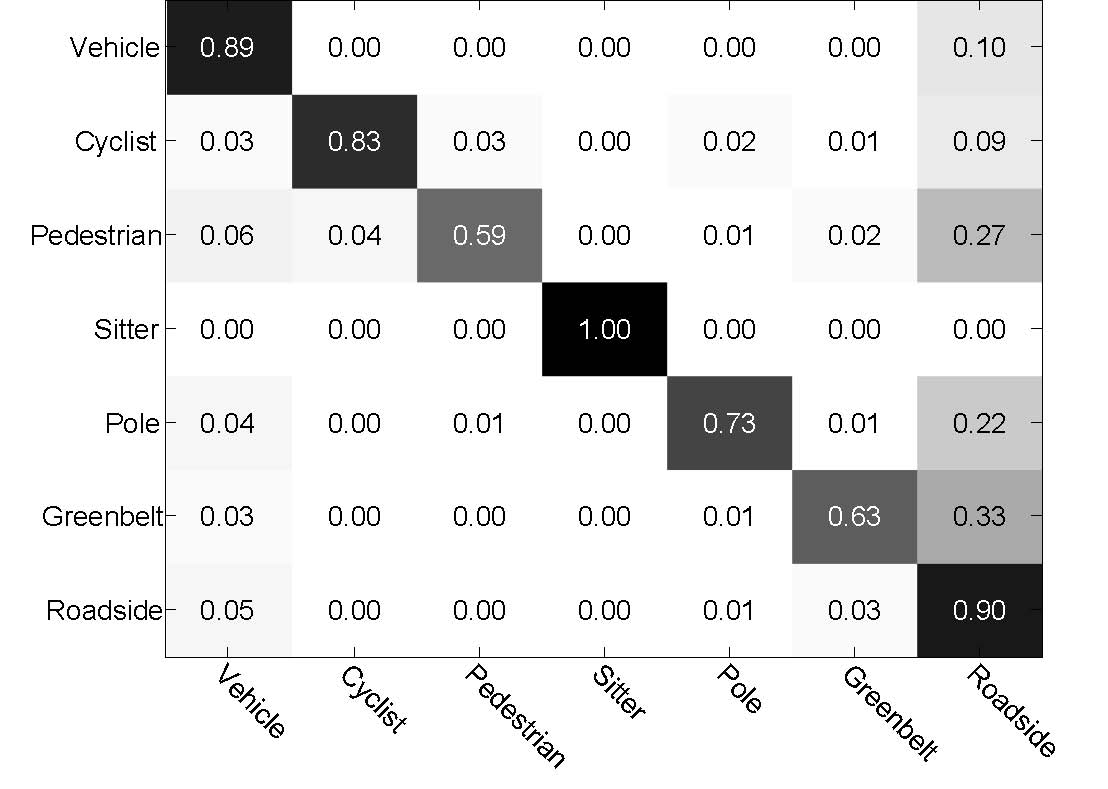}
  \centerline{(b) Unmasked RGBD}\medskip
  \end{minipage}
  \hfill
  }
    \subfigure{
  \begin{minipage}[t]{0.35\linewidth}
  \centering
   \includegraphics [width=1\textwidth]{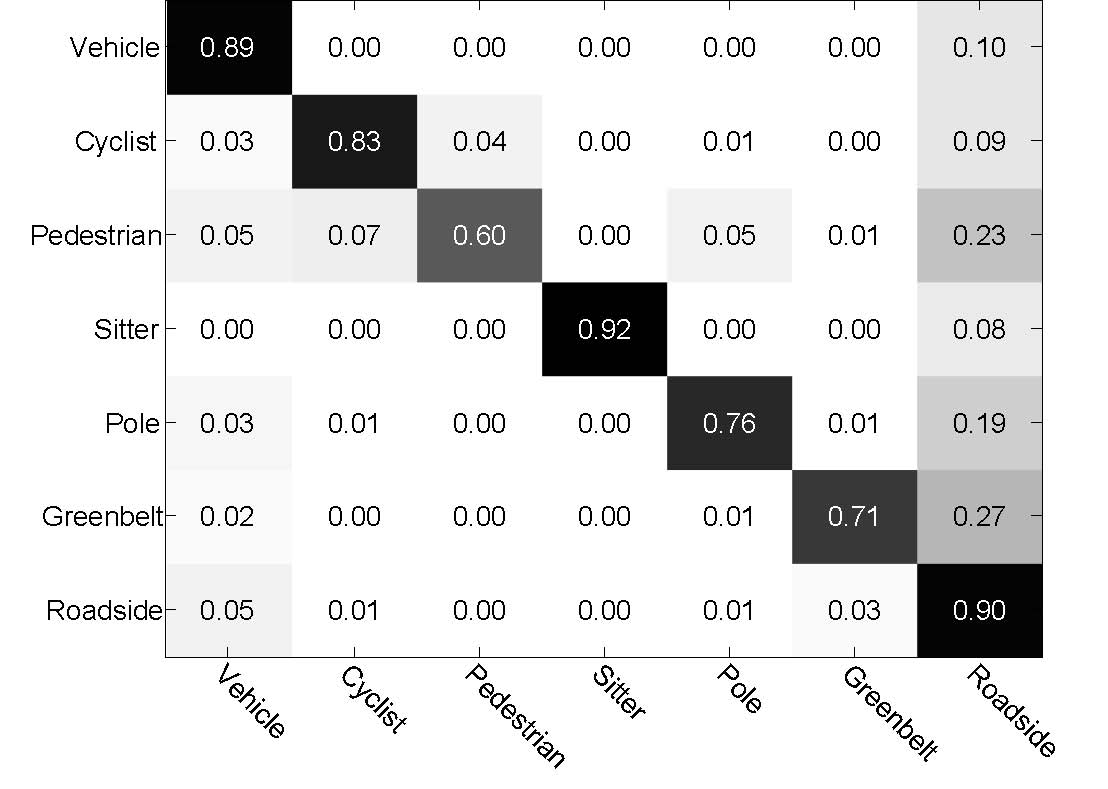}
   \centerline{(c)Masked RGB}\medskip
  \end{minipage}
  \hfill
  }
  \subfigure{
  \begin{minipage}[t]{0.35\linewidth}
   \centering
  \includegraphics [width=1\textwidth]{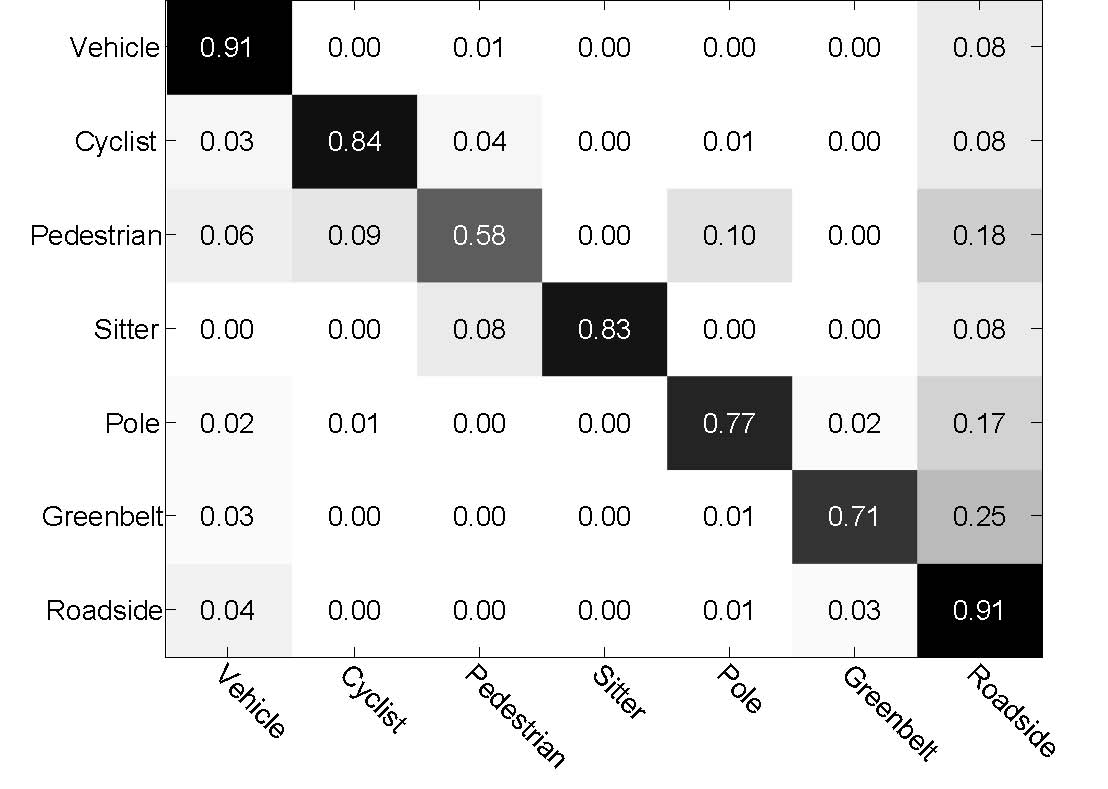}
  \centerline{(d) Masked RGBD}\medskip
  \end{minipage}
  \hfill
  }
  \vspace{-1.5em}
  \caption{The confusion matrices of different configurations.}
  \label{figure:CRNN}
\end{figure*}

In addition, we also present the confusion matrices in Fig.~\ref{figure:CRNN} to analyze the recognition performance further. These validate that the masked RGBD configuration achieves the least confusion in most of the categories. Besides this, we also make the following observations. First, among all the categories, 'Vehicle', 'Roadside', and 'Sitter' are recognized with high accuracy, followed by 'Cyclist', 'Pole', and 'Greenbelt'. The 'Pedestrian' category is most often confused. Second, we also observe that all categories are prone to be misclassified as 'Roadside'. The reason is that the 'Roadside' category is of extremely high diversity, containing variant objects such as trees, buildings, windows of the buildings, barriers on the roadside, mailboxes, and so on. Without global information, many patches of other categories are easily to be viewed as these even by human beings. Third, 'Pedestrian' is prone to be misclassified as 'Cyclist', 'Pole', or 'Roadside' due to their similarity in shape. In all, the confusions are reasonable and the CRNN performs well.

\subsection{Evaluation of Holistic Understanding}
Before evaluating the performance of holistic understanding, let us first introduce the implementation details. The parameters involved in the joint problem are empirically set as as follows. The scaling factors defined in Eq.~(\ref{eq:prob}) are $\lambda_1 = 0.5$, $\lambda_2 = 1$, $\lambda_3 = \lambda_4 = \lambda_5 = 10$; in Eq.~(\ref{eq:Fusion1}), $\alpha_0 = 1$, $\beta_0 = 500$; in Eq.~(\ref{eq:CP}), $\alpha_c = 50$; and in Eq.~(\ref{eq:Smooth1}), $\sigma=625$. Each Gaussian mixture model has five components. The algorithm is implemented in mixed Matlab/C and run on a desktop with an Intel Core i5 2300 and 12 GB memory. Our implementation has not yet been optimized for efficiency. The whole process is about $50s$ per frame. Roughly, it takes about $5s$ for loading and registering a 3D point cloud, $1s$ for point clustering, $13s$ for building the GMMs, $4s$ for the CRNN, and $22s$ for Graph Cuts inference.

Experiments are performed on the 4764 images that have not been used in the CRNN. In order to quantitatively evaluate the proposed approach, we randomly select 140 images and manually label them with both object-level segmentation and semantic category labels. When evaluating object-level segmentation, we choose the global consistency error (GCE) and the local consistency error (LCE), which are two criteria proposed by Martin et al.~\cite{Martin_2001} for measuring consistency between two segmentation results. These criteria are designed to be tolerant to different numbers of segments arising from different perceptual levels when observing complex scenarios. For semantic labeling, the average accuracy, precision, recall, and F-measure are computed.

\begin{figure*}[htbp]
\centering
\includegraphics [width=1\textwidth]{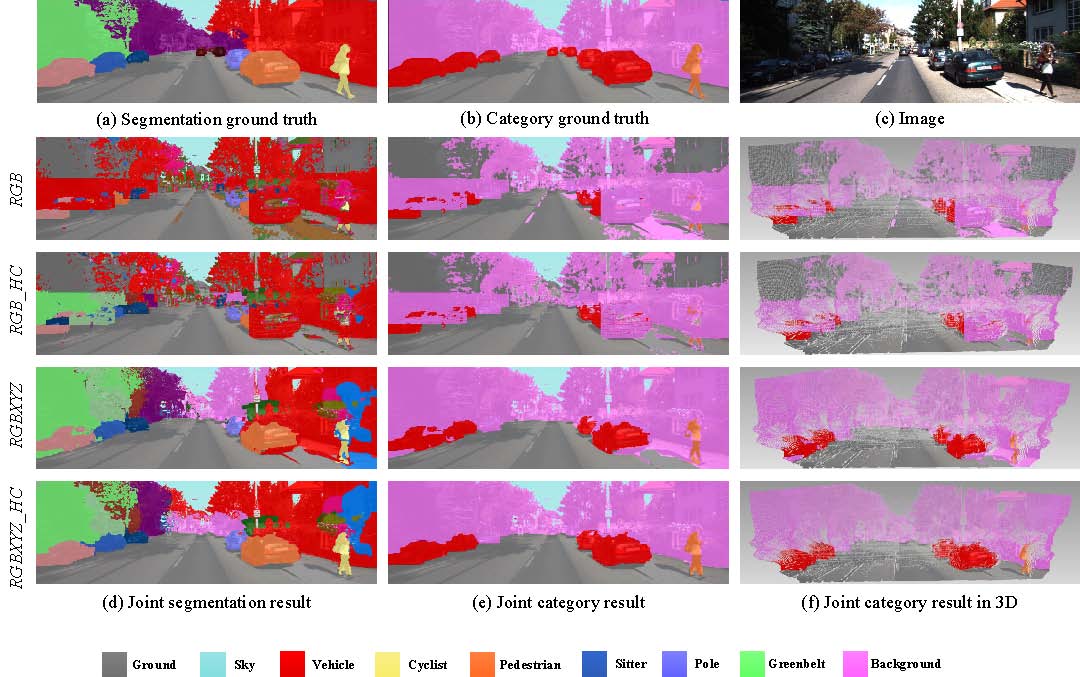}
  \vspace{-1em}
  \caption{A typical example of holistic understanding with the use of different features and constraints. Again, for segmentation results, colors have no semantic meaning.}
  \label{figure:Feature}
\end{figure*}

To investigate the performance, a group of comparative experiments is conducted. First, we are interested in how much improvement is achieved when incorporating depth information in the feature of the GMMs and integrating lidar points pivoted hard constraints (HC) into the object potential (in Sec.~\ref{sec:ObjectPotential}). According to whether location information is used and whether the HC is placed or not, we denote the algorithms by RGB, RGBXYZ, RGB\_HC, and RGBXYZ\_HC, respectively. For instance, literally, RGBXYZ\_HC represents the algorithm using both color and location features and with hard constraints, and likewise for the others. Table~\ref{tab:FeatureCompare} lists the quantitative comparison results. It shows that the incorporation of depth and hard constraints greatly improve the performance. A typical example is demonstrated in Fig.~\ref{figure:Feature}, which illustrates how these different configurations behave. From the segmentation, semantic labeling, and 3D reconstruction results in Fig.~\ref{figure:Feature}(d)(e)(f), respectively, we see that RGBXYZ\_HC outperforms the other algorithms. Note that, both 'RGB' and 'XYZ' values of the feature are all scaled to [0, 255].

\begin{table*}[htbp]\tiny\centering

\begin{tabular}{|c|c||c|c|c|c|c|}
\hline
\multicolumn{2}{|c||}{\multirow{2}*{\backslashbox{Evaluation}{Configuration}}} & Separated &\multicolumn{4}{|c|}{Holistic}\\
\cline{3-7}
\multicolumn{2}{|c||}{} & RGBXYZ\_HC &RGB & RGB\_HC & RGBXYZ & RGBXYZ\_HC\\
\hline
\hline
\multirow{2}*{Segmentation} &GCE & $0.121$ & $0.324$ & $0.187$ & $0.099$ & $\mathbf{0.090}$\\
\cline{2-7}
&LCE & $0.109$ & $0.299$ & $0.178$ & $0.094$ & $\mathbf{0.085}$\\
\hline
Category & Accuracy & $91.39\%$ & $52.71\%$ & $53.49\%$ & $91.32\%$& $\mathbf{91.97\%}$ \\
\hline
\end{tabular}
\caption{Quantitative evaluation of segmentation and semantic labeling results. Both GCE and LCE are in the range of [0, 1], where 0 signifies no error and 1 is for the worst.}
\label{tab:FeatureCompare}
\end{table*}

Finally, we investigate the performance of our holistic framework compared to the method that implements segmentation and semantic labeling separately. The quantitative comparison of object-level segmentation and average semantic labeling accuracy are listed in Table~\ref{tab:FeatureCompare} (refering to 'Separate RGBXYZ\_HC' and 'Holistic RGBXYZ\_HC'). From it we know that the holistic method achieves better performance in both segmentation and semantic labeling. To get a deeper insight, we also compare the precision and recall of each object category for semantic labeling, as listed in Table~\ref{tab:MethodsCompare}. The object categories include the seven we introduced in the CRNN, together with 'Road' and 'Sky'. The percentage of the pixels that each category holds is also listed for a reference and the total number of the pixels is $140 \times 1242 \times 375$. This table shows that both the recall and precision of 'Pedestrian', 'Pole', and 'Greenbelt' are increased in the holistic approach. Recall and precision of the other categories are either increased or decreased, which makes it difficult for us to tell the relative performance. Therefore, an F-measure that calculates the harmonic mean of the precision and recall is also provided. The F-measure of our holistic approach is improved for all categories, except 'Sky' and 'Sitter'.


\begin{table*}[htbp]\tiny\centering
 \resizebox{\textwidth}{!}{
\begin{tabular}{|c|c||c|c|c|c|c|c|c|c|c|}
\hline
\multicolumn{2}{|c||}{\backslashbox{Method}{Object Category}}& Road & Sky & Vehicle & Cyclist & Pedestrian &Sitter & Pole & Greenbelt & Roadside \\
\hline
\hline
\multicolumn{2}{|c||}{Pixel Percentage}  & $30.04$ & $5.72$ &  $9.34$ & $0.30$ & $0.15$ & $0.04$ & $0.89$ & $1.27$ & $52.25$ \\
\hline
\hline
\multirow{3}*{Separated}&Precision &  $92.37$ & $79.34$ &  $86.13$ & $81.15$ & $53.46$ & $91.11$ & $72.75$ & $35.84$ & $94.79$ \\
\cline{2-11}
&Recall &  $99.06$ & $86.84$ &  $84.48$ & $77.15$ & $36.34$ &
$66.11$ & $15.76$ & $35.40$ & $91.62$ \\
\cline{2-11}
&F-Measure &  $95.60$ & $\mathbf{82.92}$ &  $85.30$ & $79.10$ & $43.27$ & $\mathbf{76.62}$ & $25.91$ & $35.62$ & $93.18$ \\
\hline
\hline
\multirow{3}*{Holistic}&Precision &  $95.11$ & $73.68$ &  $94.32$ & $88.77$ & $64.54$ & $94.87$ & $82.17$ & $52.67$ & $93.02$ \\
\cline{2-11}
&Recall &  $97.41$ & $91.03$ &  $78.39$ & $75.06$ & $36.72$ & $64.23$ & $17.33$ & $38.55$ & $94.21$ \\
\cline{2-11}
&F-Measure &  $\mathbf{96.25}$ & $81.44$ &  $\mathbf{85.62}$ & $\mathbf{81.34}$ & $\mathbf{46.81}$ & $76.60$ & $\mathbf{28.63}$ & $\mathbf{44.52}$ & $\mathbf{93.61}$ \\
\hline
\end{tabular}}
\caption{Quantitative comparison of the proposed holistic approach versus the separated method. $\text{F-Measure} = 2 \cdot \frac{Precision \cdot Recall}{Precision + Recall}$.}
\label{tab:MethodsCompare}
\end{table*}

\begin{figure*}[thbp]\tiny
\centering
  \subfigure{
  \begin{minipage}[t]{0.9\linewidth}
  \includegraphics [width=1\textwidth]{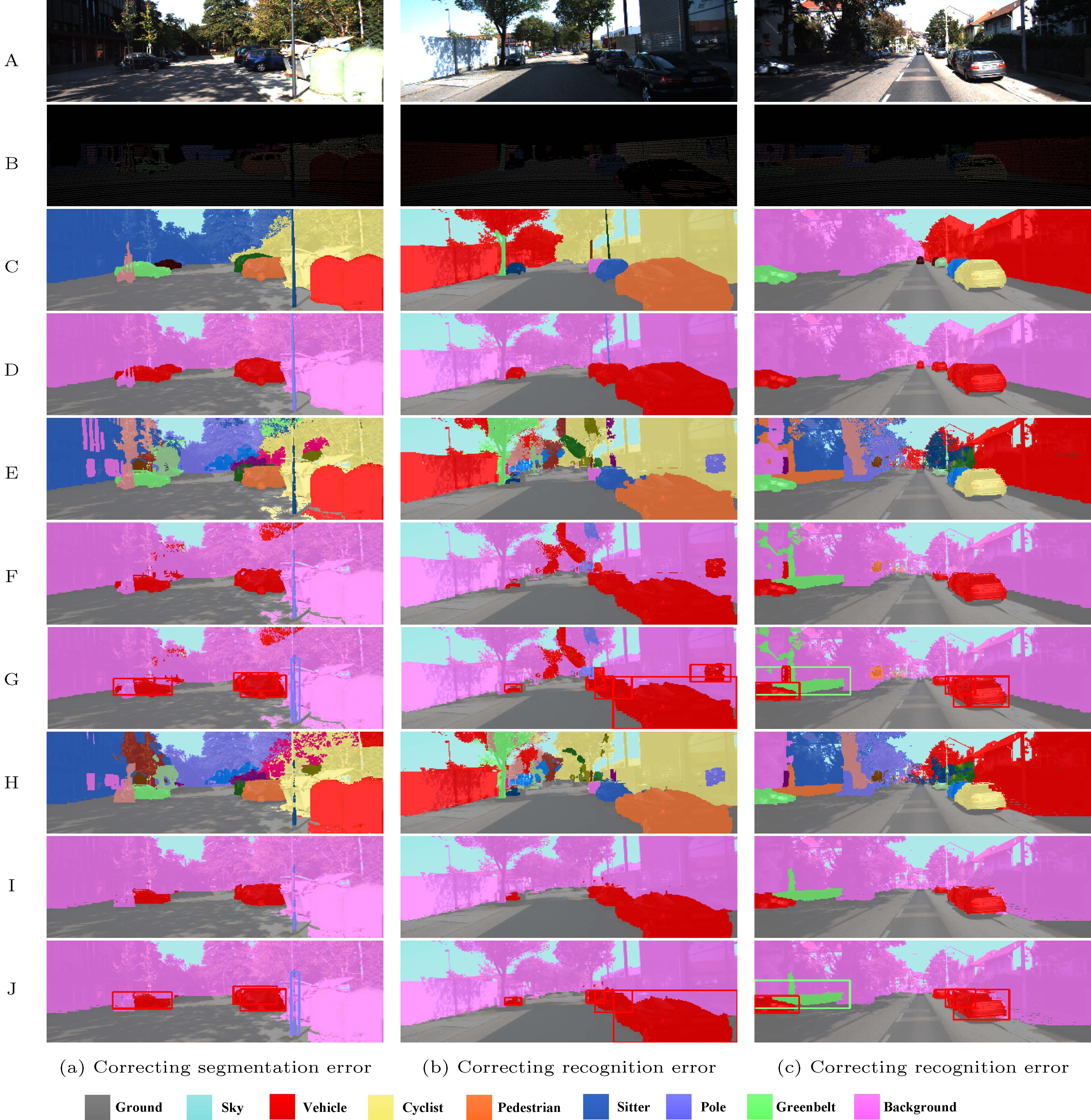}
  \end{minipage}
  }
  \vspace{-1em}
  \caption{Comparative experimental results between separated and holistic methods. Row A is images. Row B is images of produced object hypotheses. Row C and row D are the ground truth of segmentation and category labeling, respectively. Rows E-G show the segmentation, category labeling and detection result of the separated approach, and rows H-J are the three results of the holistic approach, respectively. Note that the colors on the images of segmentation have no semantic meaning. Different colors denote different objects. The colors on the images of category labeling represent the corresponding semantic categories, as shown in the legend.}
\label{fig:SeparateHolistic}
\end{figure*}

Fig.~\ref{fig:SeparateHolistic} demonstrates typical examples of how the holistic approach corrects both segmentation and semantic labeling results compared to the separated method. The improvements are presented in two aspects. On the one hand, the holistic approach can correct some segmentation errors produced by object-level segmentation. For instance, as shown in rows E to G, the separated method segments part of the roadside regions wrongly and these segmentation errors are inevitably propagated to the semantic labeling procedure. Rows H to J show that this type of errors is corrected by jointly tackling these two tasks. Such improvement benefits from the coherency considered between segmentation and semantic labeling in the holistic framework. On the other hand, the holistic approach can also correct some recognition errors of the CRNN. For example, some parts of the roadside are recognized as 'Car' and 'Pedestrian' in Fig.~\ref{fig:SeparateHolistic}(b)F-G and Fig.~\ref{fig:SeparateHolistic}(c)F-G, respectively, while with the consideration of geometrical context in our holistic framework, these recognition errors are corrected, as shown in rows I to J.

\begin{figure*}[thp]
\centering\tiny
\subfigure{
  \begin{minipage}[t]{0.9\linewidth}
  \includegraphics [width=1\textwidth]{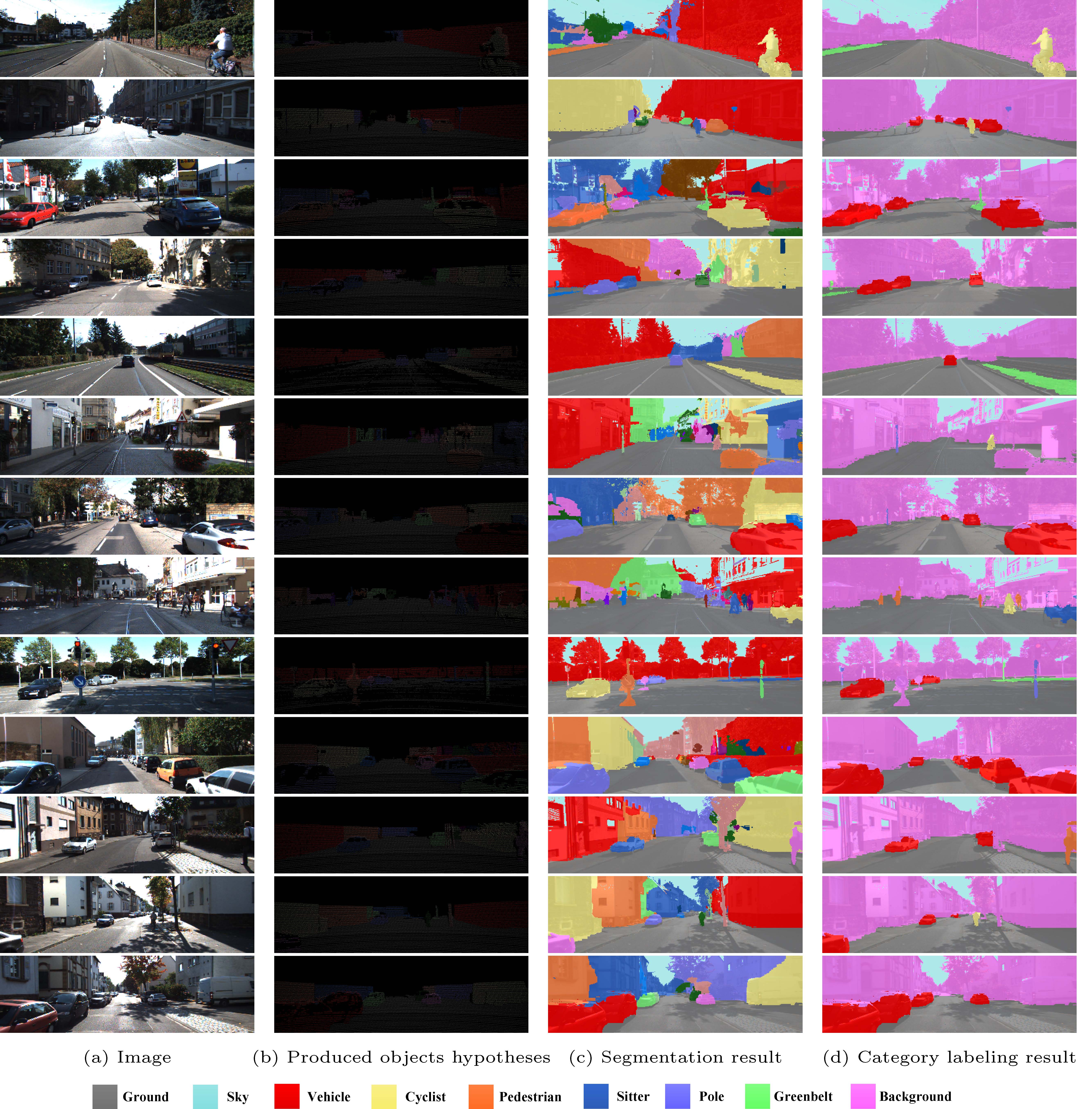}
  \end{minipage}
  }
  \vspace{-1em}
\caption{Examples of holistic scene understanding results.}
\label{fig:GoodCase}
\end{figure*}

More experimental results of the holistic approach are presented in Fig.~\ref{fig:GoodCase}. From these examples, we observe that, although the scenarios are extremely diverse, our approach can correctly segment and recognize most of the objects, such as cyclists, pedestrians, cars, poles, and backgrounds. The segmented objects are of precise boundaries.

\subsection{Discussion}
As presented above, we have conducted sets of comparative experiments. From these comparisons, we know that the integration of color and depth information highly improves the performance of both segmentation and semantic reasoning, and our holistic approach boosts the performance further. Of course, there is still room for improvement. For instance, too bright walls of buildings are easily segmented and labeled as 'Sky' and parts of cars' windows are often missed in segmentation and category labeling. These errors are mainly caused by missing lidar data. Therefore, they might be improved if the guided depth upsampling algorithm could perform better in large invalid regions.

In our experiments, we have not compared our algorithm with others' work yet. The main reason is that, although there is some object detection evaluation platform available on KITTI's website, to the best of our knowledge, there has been no work developed for object-level segmentation and semantic labeling tasks while integrating images and sparse lidar data.

\section{Conclusions and Future Work}
\label{sec:Conclusions}
In this paper, we have presented an approach for holistic road scene understanding by integrating visual and range information. The approach has been validated by extensive experiments on the challenging KITTI dataset. Both qualitative and quantitative evaluations have been performed, which show that our algorithm is promising. In future, besides improving our algorithm in the aspects discussed above, we also plan to apply this work for large scale semantic urban modeling.

\section*{Acknowledgements}
The authors would like to thank the anonymous reviewers for their helpful comments and suggestions. This research work was supported in parts by the National Natural Science Foundation of China via grants 61001171, 60534070 and 90820306, and by the Fundamental Research Funds for the Central Universities.








\end{document}